\documentclass[10pt,twocolumn,letterpaper]{article}

\usepackage{iccv}              
\usepackage{makecell}

\definecolor{iccvblue}{rgb}{0.21,0.49,0.74}
\usepackage[pagebackref,breaklinks,colorlinks,allcolors=iccvblue]{hyperref}
\usepackage[accsupp]{axessibility}
\usepackage{balance}

\def\paperID{6087} 
\def\confName{ICCV}
\def\confYear{2025}
\usepackage{multirow}
\usepackage[sort&compress]{natbib}

\title{M-Net: MRI Brain Tumor Sequential Segmentation Network via Mesh-Cast}

\author{Jiacheng Lu$^{1}$, Hui Ding$^{1*}$, Shiyu Zhang$^{1}$, Guoping Huo$^{2*}$\\
$^{1}$College of Information Engineering, Capital Normal University, 100048, China\\
$^{2}$School of Artificial Intelligence, China University of Mining and Technology-Beijing, 100083, China\\
{\tt\small jchengl@foxmail.com, dhui@cnu.edu.cn$^{*}$, sh1yuzh@163.con, kuoping@cumtb.edu.cn$^{*}$}
}

\begin{document}
\maketitle
\begin{abstract}
MRI tumor segmentation remains a critical challenge in medical imaging, where volumetric analysis faces unique computational demands due to the complexity of 3D data. The spatially sequential arrangement of adjacent MRI slices provides valuable information that enhances segmentation continuity and accuracy, yet this characteristic remains underutilized in many existing models. The spatial correlations between adjacent MRI slices can be regarded as “temporal-like” data, similar to frame sequences in video segmentation tasks. To bridge this gap, we propose M-Net, a flexible framework specifically designed for sequential image segmentation. M-Net introduces the novel Mesh-Cast mechanism, which seamlessly integrates arbitrary sequential models into the processing of both channel and temporal information, thereby systematically capturing the inherent “temporal-like” spatial correlations between MRI slices. Additionally, we define an MRI sequential input pattern and design a Two-Phase Sequential (TPS) training strategy, which first focuses on learning common patterns across sequences before refining slice-specific feature extraction. This approach leverages temporal modeling techniques to preserve volumetric contextual information while avoiding the high computational cost of full 3D convolutions, thereby enhancing the generalizability and robustness of M-Net in sequential segmentation tasks. Experiments on the BraTS2019 and BraTS2023 datasets demonstrate that M-Net outperforms existing methods across all key metrics, establishing itself as a robust solution for temporally-aware MRI tumor segmentation.

\end{abstract}

\begin{figure}[h]
	\centering
	\includegraphics[width=0.5\textwidth]{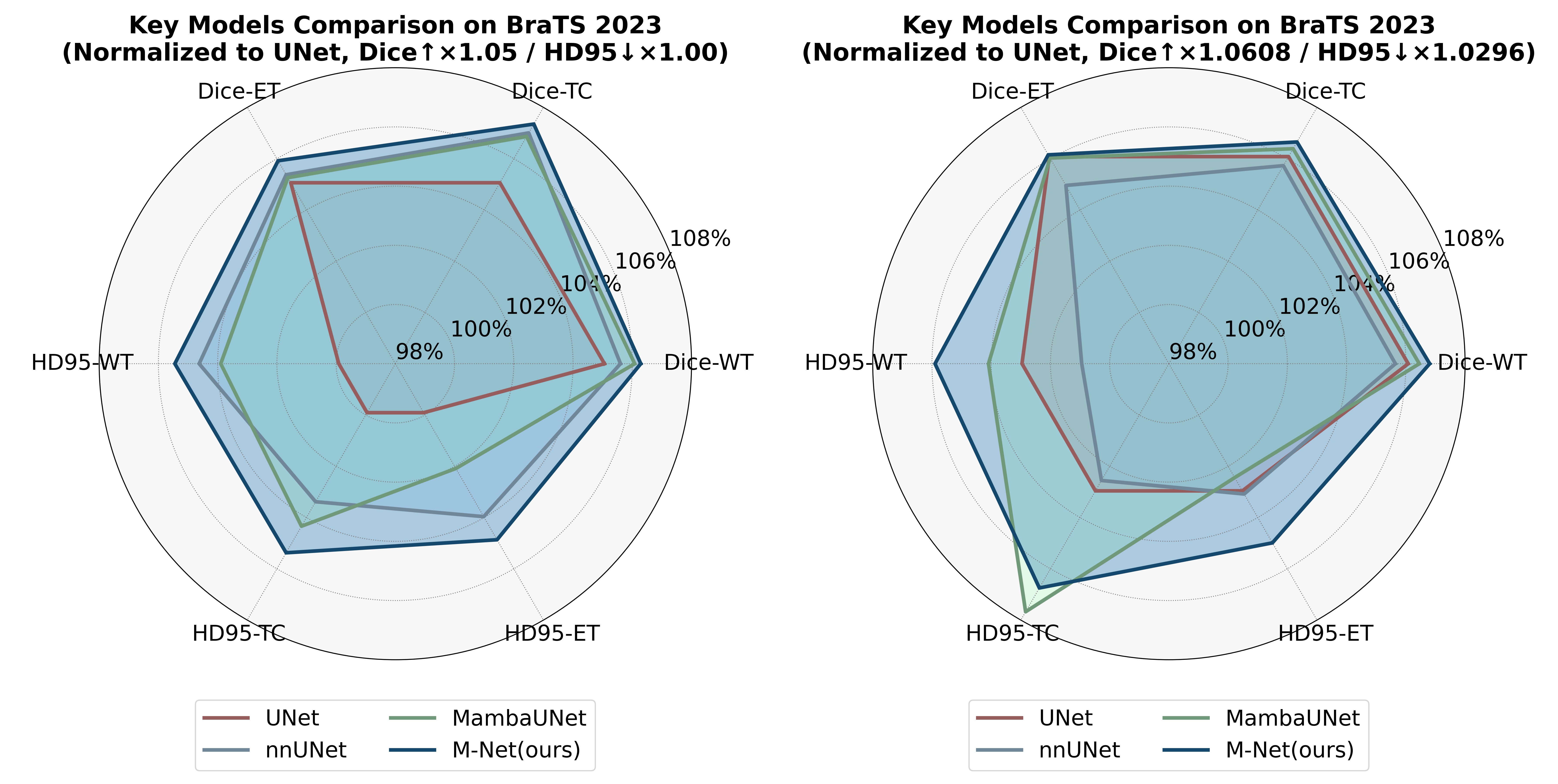} 
	\caption{
		Performance radar charts of M-Net and several mainstream models on BraTS 2023/2019. The values in the charts are rescaled, with larger values indicating better performance.}
	\label{fig:com_res_figure}
\end{figure}

\begin{figure}[ht]
	\centering
	\includegraphics[width=0.45\textwidth]{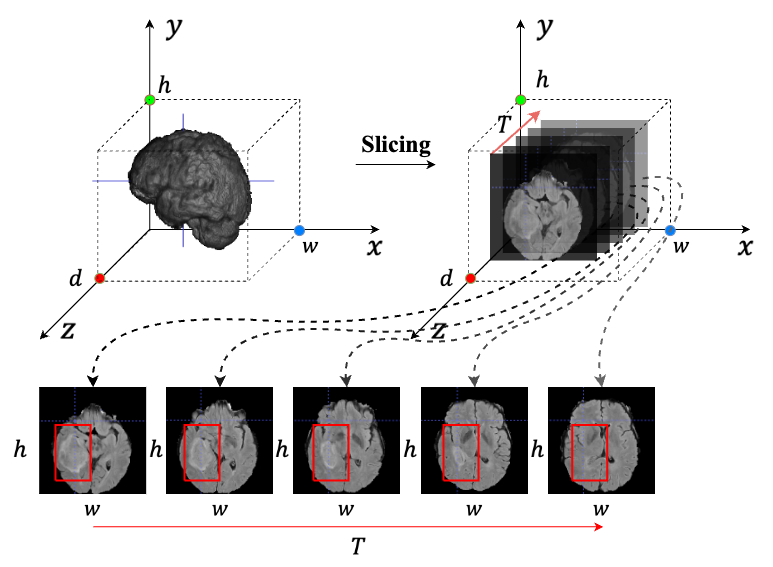} 
	\caption{“Temporal-like” spatial correlations in MRI. For an MRI slice sequence, the position and size of the lesion change with spatial continuity through the sequence of slices.}
	\label{fig:intro}
\end{figure}

\section{Introduction}
\label{sec:intro}
Accurate brain tumor segmentation is essential for disease diagnosis and treatment planning in medical imaging\cite{Tan1, Liang1}. However, brain tumor MRI images pose significant challenges due to irregular tumor boundaries, varying locations, complex textures, inconsistent grayscale levels, and low interclass contrast. In recent years, deep learning\cite{Fuku} has achieved remarkable results in medical image segmentation. A key milestone was the UNet\cite{Ron1}, a 2015 encoder-decoder segmentation network proposed by O. Ronneberger et al. Many subsequent studies have introduced improvements, such as CANet\cite{canet} and MIRAU-Net\cite{mirau} with convolutional attention, UKAN\cite{UKAN} with knowledge-aware networks (KAN)\cite{kan},  and Med-SAM\cite{MedSAM}, which improved generalizability. Progress in natural language processing has influenced medical image segmentation, with adaptations of text models like LSTM(Long Short-Term Memory)\cite{lstm}, Transformer\cite{transformer}, and Mamba SSM(State Space Model)\cite{mamba} gaining popularity. Hybrid architectures, such as TransUNet\cite{transunet}, TransNorm\cite{transnorm}, MedNeXt\cite{mednext}, UNETR\cite{unetr}, and Swin UNETR\cite{swin}, combine UNet with Transformer-based designs, while classic RNN models like LSTM\cite{lstm} have found some applications in medical image segmentation\cite{Han1, Xu1, Dan1, Sha1}. Additionally, Mamba SSM\cite{mamba} and its visual module, VMamba\cite{liu2024vmamba}, have shown excellent performance, leading to algorithms like Mamba UNet\cite{mambaunet} gaining prominence in the field.

\begin{figure}[ht]
	\centering
	\includegraphics[width=0.43\textwidth]{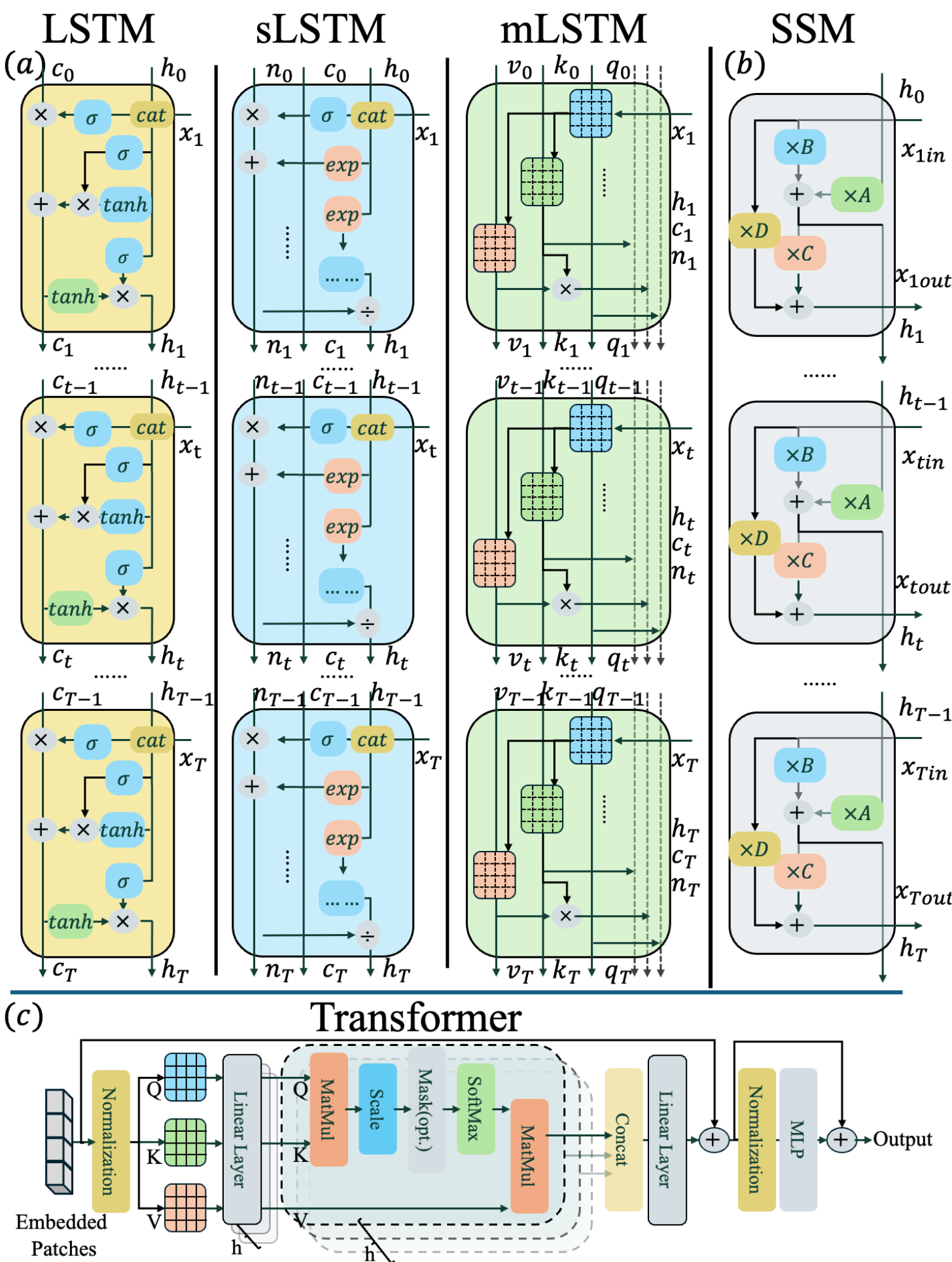} 
	\caption{Categories of Mainstream Sequence Models: (a) the LSTM series, including LSTM\cite{lstm}, sLSTM\cite{xlstm}, and mLSTM\cite{xlstm}; (b) Mamba SSM\cite{mamba}; and (c) Transformer\cite{transformer}.}
	\label{fig:related}
\end{figure}

While segmentation algorithms with language-based sequential modules have advanced 2D and 3D segmentation\cite{Mehta, Beers, Feng, Zhang1}, their sequence modules are generally limited to patch sequences within single images. Compared to other modalities, brain tumor MRI slice data exhibit spatial correlations between adjacent slices, reflected in variations in lesion size and location, as illustrated in Figure.~\ref{fig:intro}. The presence of these spatial correlations, along with the continuous growth patterns of human organs and lesions, can be regarded as “temporal-like” data, analogous to those in video segmentation tasks\cite{video1, video2}.

Although such “temporal-like” data can be learned using 3D models, 2D slice-based strategies are more prevalent in practical applications and deployment scenarios due to their better adaptability to limited computational resources. Notably, the “temporal-like” spatial correlations are challenging to directly observe in 2D slice-based tasks. This limitation may partially explain why 2D slice-based algorithms generally underperform compared to 3D models.

To overcome the limitations of current MRI segmentation algorithms, particularly 2D slice-based methods, this study breaks through the constraints of traditional 2D and 3D algorithms by defining a “temporal-like” MRI modeling approach. We further propose a modular sequential image segmentation framework—M-Net. This framework leverages the designed Mesh-Cast mechanism in combination with any sequential processing algorithm to capture sequential correlations across spatial, channel, and temporal dimensions. While maintaining the computational efficiency of 2D algorithms, it extracts 3D volumetric contextual “temporal-like” information. The key contributions of this study are as follows:

\begin{itemize}
    \item We reconceptualize MRI segmentation by treating MRI slice sequences as "temporal-like" data. Through our novel Mesh-Cast mechanism, M-Net seamlessly integrates with diverse temporal algorithms (RNN, LSTM, SSM, Transformer), establishing a unified framework for sequence-based image segmentation.
    \item The Mesh-Cast mechanism and its Sequential Module are designed to propagate sequential images into temporal algorithms. This mechanism enables traditional 2D models to capture information across both the temporal and feature channel dimensions, enhancing their capability to process multi-dimensional data.
    \item A Two-Phase Sequential (TPS) training strategy is proposed, which systematically varies sequence order during training. This approach enables models to first learn common patterns across sequences before adapting to sequence-specific features, thereby improving both generalization ability and model robustness.

\end{itemize}
The results on BraTS 2019 and 2023 demonstrate that the proposed algorithm achieves state-of-the-art performance
while maintaining a low computational time (see Figure.~\ref{fig:com_res_figure}).

\section{Related Work}

\begin{figure*}[ht]
	\centering
	\includegraphics[width=0.9\textwidth]{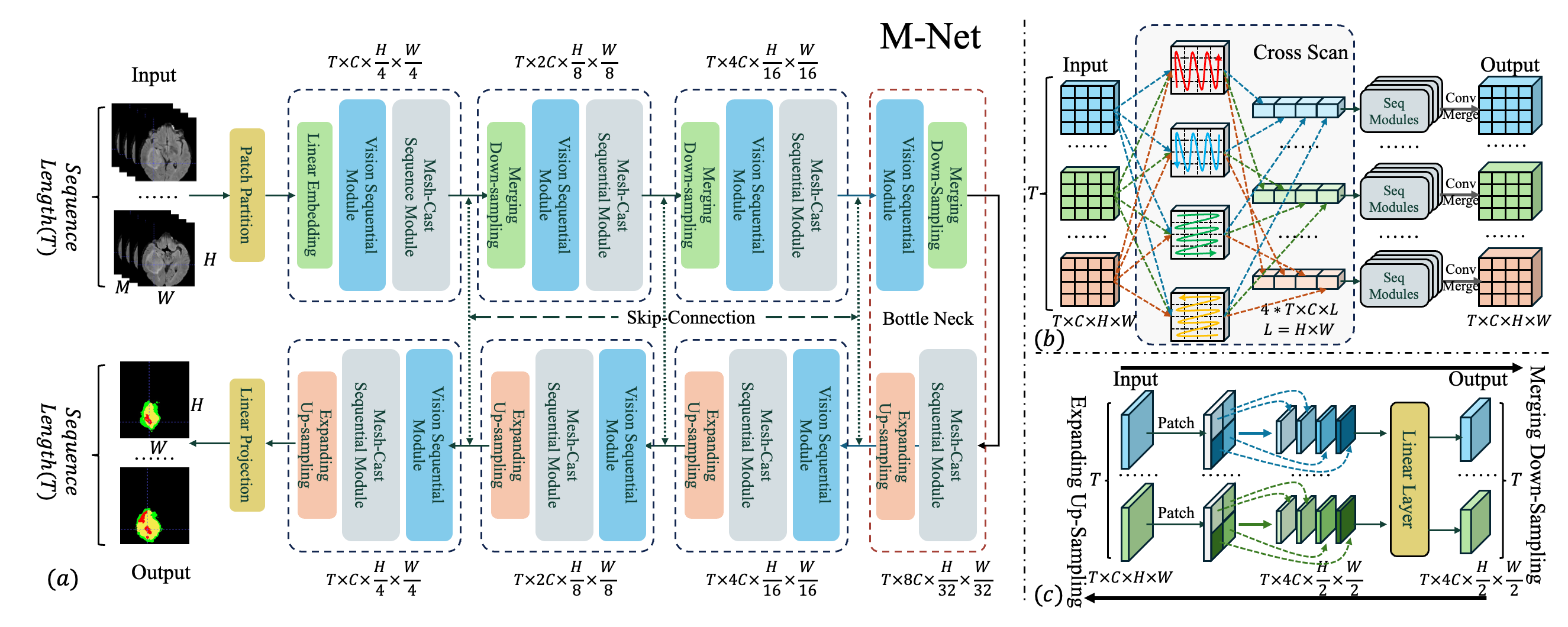} 
	\caption{The overall structure of M-Net. (a) An encoder-decoder framework of M-Net, (b) the integrated Vision Module, (c) the up/down-sampling module, and the novel Mesh-Cast Sequential Module.}
	\label{fig:mnet}
\end{figure*}

With advancements in natural language processing, numerous novel sequential processing algorithms have emerged, as shown in Figure~\ref{fig:related}. These algorithms are primarily applied in the visual domain and are widely used in image sequence processing tasks such as video analysis, owing to their unique capability in handling sequential information. For the “temporal-like” correlations present in MRI slices, sequential models can be further specialized to enhance the representation of these relationships.

Besides the widely used Transformer (Figure~\ref{fig:related}(c)), the LSTM family (Figure~\ref{fig:related}(a)) also shows great potential. ConvLSTM\cite{Pfe} uses convolution in place of standard LSTM units to capture temporal features. The recent xLSTM\cite{xlstm} combines sLSTM and mLSTM with residual connections to improve feature extraction and parallelism. sLSTM introduces Exponential Gating and Normalization for better stability and accuracy in long sequences, while mLSTM upgrades vector operations to matrix form, enhancing memory and parallel capacity.

Additionally, the State Space Model (SSM) family’s Mamba SSM (as shown in Figure~\ref{fig:related}(b))\cite{mamba} has recently demonstrated strong performance in both text and visual domains, leading to the development of efficient algorithms such as Mamba UNet, which has been applied to medical image segmentation tasks. Its core mechanism, the Selective Scan Space State Sequential Model (S6, Selective Scan Model), is a learnable SSM structure that updates hidden layer parameters using an RNN framework.

Each of these sequential models has unique characteristics and can be explored for capturing “temporal-like” features in MRI slices. However, they are more commonly used for sequence block processing within single images. Since the limitations of “temporal-like” data have not been sufficiently addressed, existing research has rarely investigated the use of sequential algorithms to capture 3D volumetric contextual correlations in slice-based data.

\section{Method}
\begin{figure*}[ht]
	\centering
	\includegraphics[width=0.8\textwidth]{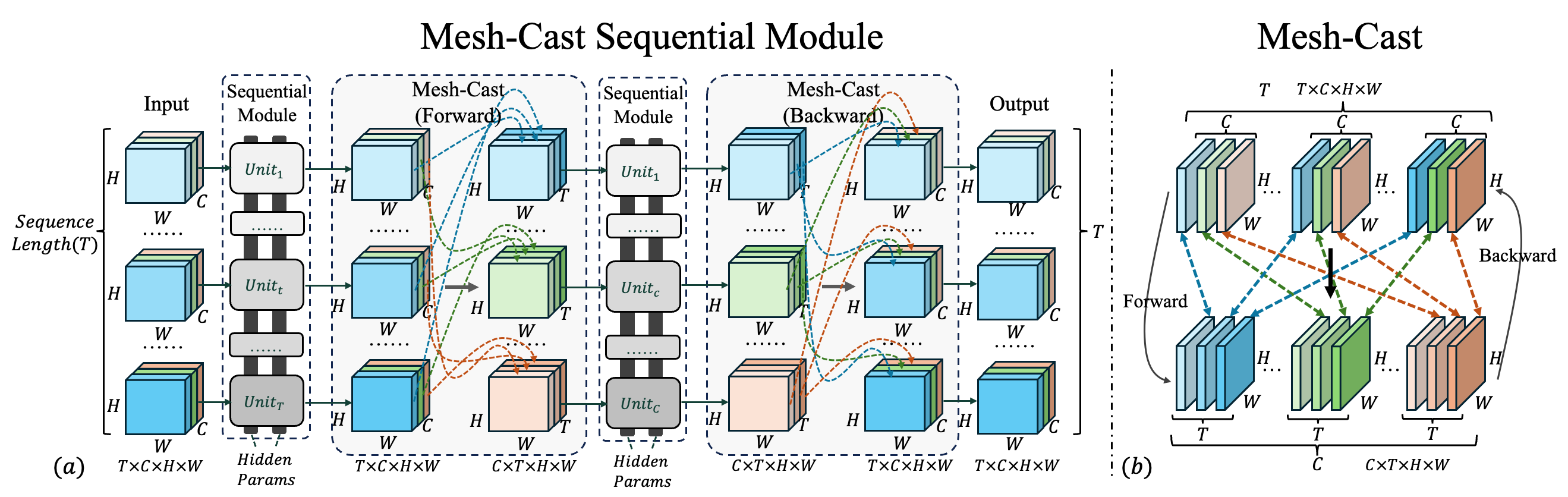} 
	\caption{Diagram of the Mesh-Cast Sequential Module. (a) Integration of the Mesh-Cast mechanism with sequential module units, (b) the Mesh-Cast mechanism adjusts the focus dimension of the feature sequence.}
	\label{fig:meshcast}
\end{figure*}

Given the strong structural consistency and minimal scale variation in brain MRI sequences, capturing these temporal characteristics using a Sequential Module presents a promising research direction. To this end, we propose the M-Net(Mesh-Cast Net, Figure~\ref{fig:mnet}) framework for sequential image segmentation. Through the Mesh-Cast mechanism(Figure~\ref{fig:meshcast}) we defined, this framework can flexibly select and utilize any sequential data processing algorithms to capture feature correlations across the sequence and channel dimensions, including LSTM series, Transformer, and Mamba SSM(Figure~\ref{fig:mamba}). Additionally, we define a sequence input mode distinct from traditional slice-based input and design the TPS (Two-Phase Sequential, Figure~\ref{fig:tps}) training strategy to help sequential models better capture the “temporal-like” correlations in MRI sequences.

On this basis, we designed the Two-Phase Sequential (TPS) training strategy for training processes in sequential image segmentation algorithms.

\subsection{M-Net Framework}
The M-Net framework incorporates a classic encoder-decoder structure and a Skip-Connection structure, (as shown in Figure.~\ref{fig:mnet}(a)) and uses a Patch-based up/down-sampling module in each encoder-decoder layer (Figure.~\ref{fig:mnet}(c)). Each layer contains both the Vision Sequential Module (Figure.~\ref{fig:mnet}(b)) and the Mesh-Cast Sequential Module (Figure.~\ref{fig:meshcast}), which are responsible for capturing the temporal correlations between the 2D images, image feature channels, and image sequences, respectively. Both the Vision Sequential Module and the Mesh-Cast Sequential Module include a Sequential Module interface, which can freely replace any temporal model mentioned in Section 2.

Through the proposed Mesh-Cast mechanism and its internal Sequential Module, M-Net can accept and process multimodal images with sequential characteristics. Its input is a set of multimodal MRI sequences:
\begin{align}
X = \{x_1, x_2, \dots, x_T\}
\end{align}
where \(x_t \in \mathbb{R}^{H \times W \times C}\) represents a single multimodal MRI slice, \(t \in [1, T]\) denotes the frame number in the slice sequence, i.e., the sequence length, and \(H\), \(W\), and \(C\) represent the height, width, and the number of channels in the slice image, respectively.

For the input image sequence \(X\), M-Net first performs image-level feature extraction through the Vision Sequential Module. The Vision Sequential Module utilizes a Cross-Scan-based scanning approach to serialize the input images. For each slice \(x_t \in \mathbb{R}^{H \times W \times C}\) in the sequence \(X\), the scanning process is as follows:
\begin{align}
x^{\text{LR}}_{i,j} = [x_{i,1}, x_{i,2}, \cdots, x_{i,j}] \\
x^{\text{RL}}_{i,j} = [x_{i,j}, x_{i,j+1}, \cdots, x_{i,w}] \\
x^{\text{TB}}_{i,j} = [x_{1,j}, x_{2,j}, \cdots, x_{i,j}] \\
x^{\text{BT}}_{i,j} = [x_{i,j}, x_{i+1,j}, \cdots, x_{h,j}]
\end{align}
Cross-Scan consists of four scanning directions: left to right, right to left, top to bottom, and bottom to top. Where\(\quad j = 1, 2, \dots, w \), \(\quad i = 1, 2, \dots, h\).  \(x_{i,j}\) represents the feature value at the position of the \(i\)-th row and \(j\)-th column. Subsequently, the four directional feature sequences are combined:
\begin{align}
x^{\text{CS}}_{i,j} = [x^{\text{LR}}_{i,j}, x^{\text{RL}}_{i,j}, x^{\text{TB}}_{i,j}, x^{\text{BT}}_{i,j}]
\end{align}
This approach allows the extraction of sequential correlations in four directions, ensuring that the sequence module maximizes understanding of relationships between 2D image positions and features within the sequence. Then, we propose the Mesh-Cast Sequential Module to capture temporal and channel-wise sequential correlations.

\subsection{Mesh-Cast Sequential Module}

The Mesh-Cast Sequential Module is the core component of M-Net. It treats multimodal MRI slices as temporal sequences and enables the sequence model to perceive and capture temporal and channel (modality) correlations by repeatedly exchanging the temporal and channel dimensions through a mesh-based propagation mechanism. The algorithmic flow of this module is illustrated in Figure~\ref{fig:meshcast}, which primarily consists of two sets of temporal modeling modules along with the forward and backward propagation of the Mesh-Cast process.

\begin{figure}[ht]
	\centering
	\includegraphics[width=0.4\textwidth]{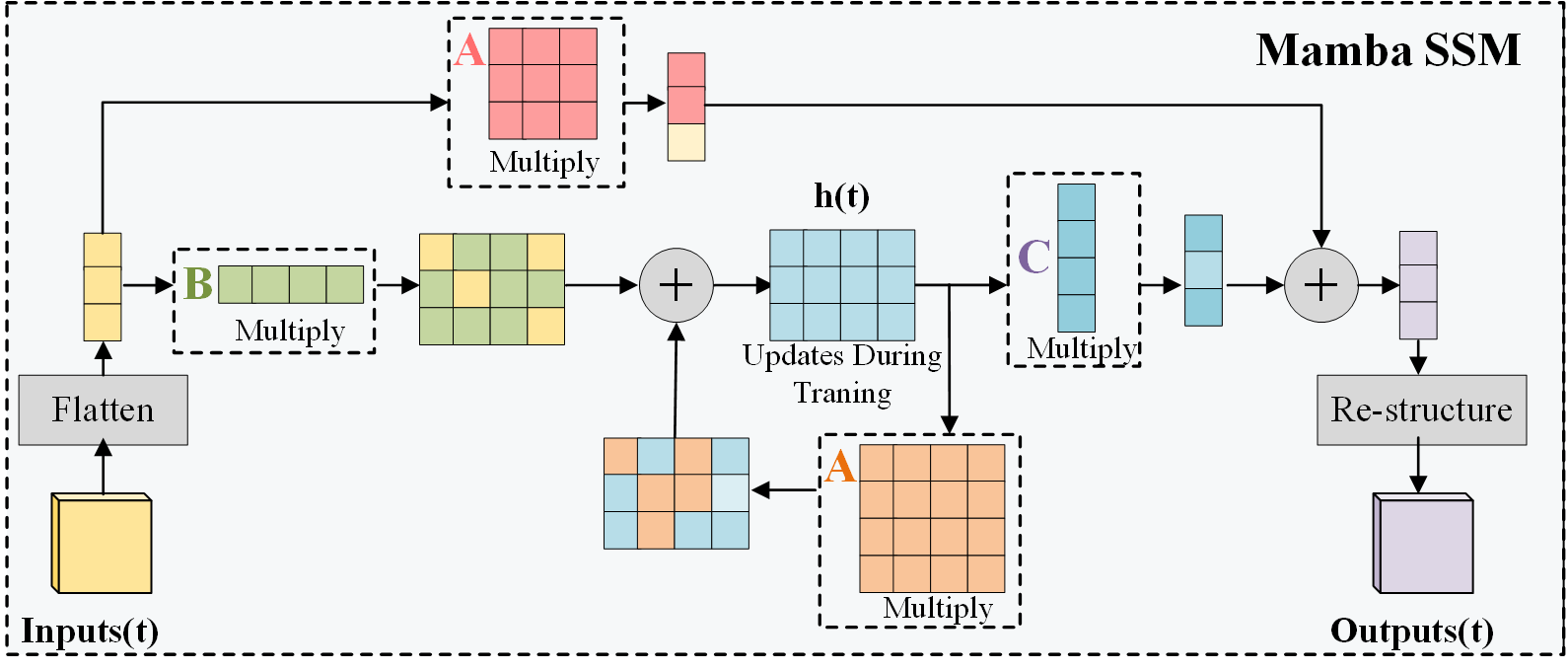} 
	\caption{
		Diagram of the structures for Mamba SSM. These matrix parameters \( \mathbf{A} \), \( \mathbf{B} \), \( \mathbf{C} \), and \( \mathbf{D} \) are learned through training.}
	\label{fig:mamba}
\end{figure}

For a given feature sequence \( X_{\text{in}} \), the Mesh-Cast Sequential Module first performs sequence perception in the temporal dimension. The temporal frame number \( T \) of the input \( X_{\text{in}} \) serves as the sequence length for the Sequential Module. Here, \( X_{\text{in}} \) can be represented as:
\begin{align}
X_{\text{in}} = \{x_1, x_2, \dots, x_T\}, \quad x_t \in \mathbb{R}^{H \times W \times C}
\end{align}
where \( T \) is the number of frames, i.e., the temporal length, \( H \) and \( W \) are the height and width of each frame image, and \( C \) is the number of channels (e.g., the different signal channels in MRI images). Subsequently, this 2D feature sequence is flattened into a 1D temporal sequence:
\begin{align}
X_{\text{seq}} = \{x_1, x_2, \dots, x_T\}, \quad x_t \in \mathbb{R}^{C \times D}
\end{align}
where \( T \) remains the temporal length of \( X_{\text{in}} \), and \( D \) is the feature dimension obtained by flattening the slice, i.e., \( D = H \times W \). The Sequential Module will focus on the sequential correlation within the temporal length \( T \), where the channel dimension \( C \) of each \( x_t \) will be set as a parameter similar to the batch size in Mesh-Cast, and will not participate in the calculation. Then, \( X_{\text{seq}} \) is input into the Sequential Module:
\begin{align}
x_{\text{seq}_t}' = \text{Sequential Unit}_t (x_{\text{seq}_t}), \quad x_{\text{seq}_t}' \in \mathbb{R}^{C \times D}
\end{align}
where \( \text{Sequential Unit}_t \) refers to each temporal processing unit in the Sequential Module, and the number of units equals the temporal length of the input \( X_{\text{seq}} \). The Sequential Module can be freely replaced by temporal algorithms such as LSTM, ConvLSTM, Transformer, xLSTM and Mamba SSM(as shown in Figure.~\ref{fig:mamba}). The RNN series algorithms, due to their temporal modeling capabilities, are more suitable for processing image sequences.

After processing the \( T \) frames, the Sequential Module generates an output feature sequence with the same size as the input: \(\quad X_{\text{seq}}' \in \mathbb{R}^{C \times T \times D}\) . Next, the sequence \( X_{\text{seq}}' \) undergoes dimensional reorganization via Mesh-Cast (Forward). 
\begin{align}
    X_{\text{channel}} = \text{Transpose}_{\text{forward}}(X_{\text{seq}}', (0, 1))
\end{align}
Specifically, the channel dimension \( C \) in \( X_{\text{seq}}' \) is treated as the sequence length, and the temporal frame number \( T \) is set as a parameter similar to batch size, temporarily ignored in subsequent calculations. The reorganized sequence can be represented as:
\begin{align}
X_{\text{channel}} = \{x_1, x_2, \dots, x_C\}, \quad x_c \in \mathbb{R}^{T \times D}
\end{align}
After Mesh-Cast (Forward), \( X_{\text{channel}} \) is input into a Sequential Module with the number of units equal to the number of channels in \( X_{\text{channel}} \):
\begin{align}
x_{\text{channel}_c}' = \text{Sequential Unit}_c (x_{\text{channel}_c})
\end{align}
Finally, the output \(X_{\text{channel}}' \in \mathbb{R}^{T \times C \times D} \), after processing in the Sequential Module for both temporal and channel dimensions, will be recombined to its initial dimensions and size through Mesh-Cast (Backward) and used as input to other modules:
\begin{align}
X_{\text{out}} = \text{Transpose}_{\text{backward}}(X_{\text{channel}}', (0, 1))\\
X_{\text{out}} = \{x_1, x_2, \dots, x_T\}, \quad x_t \in \mathbb{R}^{H \times W \times C}
\end{align}
Since many Sequential Modules have the property of being stackable, M-Net recommends applying a Layer-Attention mechanism when using multiple layers of Mesh-Cast Sequential Modules. Specifically, for \( n \) stacked layers with outputs \( Y_i, i \in [1, n] \), a SE (Squeeze-and-Excitation) layer-level attention is used to compute the weights \( \alpha_i \) of each layer. The features of each layer are weighted and residual connections are made with the original input:
\begin{align}
Y_{\text{final}} = X_{\text{input}} \cdot Y_{\text{balanced}} + \sum_{i=2}^n \beta_i \cdot Y_i
\end{align}
where \( \beta_i \) is the weighting coefficient for the auxiliary layer features, typically set to a value less than 1 to avoid over-amplifying non-primary features.

\subsection{TPS Training Strategy}
\begin{figure}[h]
	\centering
	\includegraphics[width=0.42\textwidth]{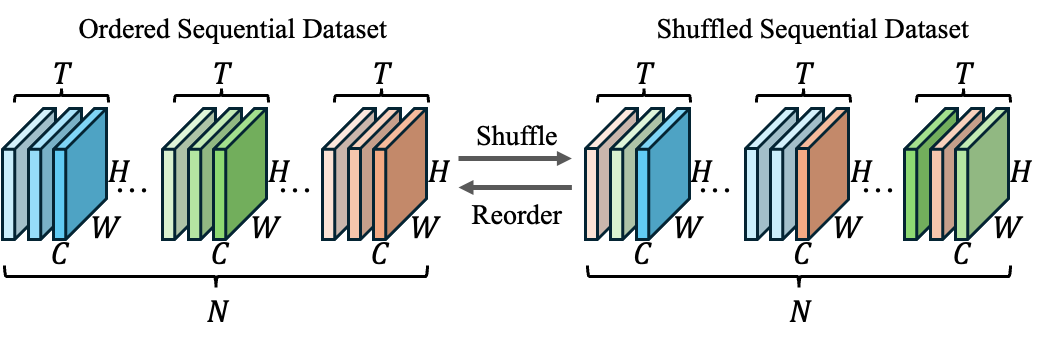} 
	\caption{
		Illustration of the TPS training strategy. In the first phase, the model is trained on the Shuffled Sequential Dataset, while in the second phase, it returns to the Ordered Sequential Dataset for fine-tuning. }
	\label{fig:tps}
\end{figure}

Considering the model structure of M-Net, the input data, and their differences from traditional 2D or 3D segmentation models, we propose a training strategy for sequence models and sequential data: the TPS (Two-Phase Sequential) training strategy. TPS is a two-phase training strategy aimed at enhancing the generalization ability of the model, as shown in Figure.~\ref{fig:tps}.

Let the initial input \( \mathcal{X} \) be a dataset composed of \( n \) sequences \( X_i \), where each sequence \( X_i \) contains \( t \) elements. The initial input dataset and each sample sequence can be represented as:
\begin{align}
\mathcal{X} = \{X_1, X_2, \dots, X_n\}, \quad X_i \in \mathbb{R}^{T \times H \times W \times C}\\
X_i = (x_{i1}, x_{i2}, \dots, x_{iT}), \quad x_{it} \in \mathbb{R}^{H \times W \times C}
\end{align}
In the first phase, TPS performs a frame-level Shuffle operation on the input sequence sample dataset, generating a new shuffled dataset \( \mathcal{X}' \), whose elements can come from any sequence \( X_i \) and any position \( x_{it} \):
\begin{align}
\mathcal{X}' = \{X_1', X_2', \dots, X_n'\}, \quad X_i' = (x_{i1}', x_{i2}', \dots, x_{iT}')
\end{align}
Here, \( x_{it}' \) represents any shuffled sequence element from the original dataset. These sequence elements will be globally shuffled, meaning that each new sequence element may come from any position in other sequences. By training the model with the shuffled dataset \( \mathcal{X}' \) as input, the model can better adapt to various input patterns and learn a broader range of features. At the same time, the diverse sequence inputs will accelerate the model's convergence.

In the second phase, TPS restores the input sequence samples so that the model can fine-tune on samples with real sequence correlations. Compared to the first phase, the data diversity in the second phase will decrease, but the authenticity and fine-grained sequence correlations will provide the model with further convergence ability.

For LSTM-based models, the diversified samples in the TPS training strategy will help in modeling long-distance understanding of common features in lesions. Real sequence data will also help overcome the erroneous information introduced during the first phase. The global generalization ability of Mamba SSM will also lead to an improved understanding of lesion location and size during the two phases of TPS training.

\subsection{Loss Function}
The multiclass brain tumor segmentation task can be converted into a multi-channel single-class segmentation task, using a combined loss function of BCE Loss and Dice Loss\cite{diceloss} during training. BCE Loss is defined as:
\begin{align}
L_{\text{BCE}} = -\sum_{i=1}^{W} \sum_{j=1}^{H} \left[ T_{ij} \log(P_{ij}) + (1 - T_{ij}) \log(1 - P_{ij}) \right]
\end{align}
where \( W \) and \( H \) denote the width and height of the predicted image \( P(i,j) \) and the ground truth \( T(i,j) \). As cross-entropy loss calculates pixel-wise errors, it may be affected by the imbalance between positive (lesion) and negative (normal) samples in images. The Dice Loss function, specifically designed for handling class imbalance, provides a more accurate reflection of segmentation performance on small lesion regions in medical image segmentation:
\begin{align}
L_{\text{Dice}}(P, T) = 1 - \frac{2 \times (\sum_{i=1}^{N} p_i t_i + \tau)}{\sum_{i=1}^{N} p_i + \sum_{i=1}^{N} t_i + \tau}
\end{align}
where \( p_i \in P \) represents the predicted image, \( t_i \in T \) represents the ground truth, \( \tau \) is a small constant, and \( N \) is the total number of pixels. Based on these loss functions, the combined joint loss function is formulated as follows:
\begin{align}
L_{\text{joint}} = \sum_{i=1}^{3} \left( \lambda L_{\text{Dice}_i} + (1 - \lambda) L_{\text{BCE}_i} \right)
\end{align}
where \( \lambda \) is a weighting factor in the range \( 0 < \lambda < 1 \) that balances the different losses. To reduce task complexity, we divide the segmentation task into a single-class segmentation with multiple channels, calculating losses separately. Here, \( L_{\text{dice}_i} \) and \( L_{\text{bce}_i} \) denote the Dice Loss and BCE Loss for the \( i \)-th channel, respectively.

\section{Experiments}
\subsection{Datasets}
The BraTS-2019 and BraTS-2023 datasets\cite{data1, data2, data3}, released by MICCAI in 2019 and 2023, respectively, include brain tumor data. The training sets contain 335 and 1251 cases, respectively, with each MRI image sized \(155 \times 240 \times 240\). We split each dataset into training and testing sets at an 8:2 ratio. Each subject’s data includes four MRI modalities (T1, T1c, T2, and FLAIR) and four labels (0 for non-tumor, 1 for necrotic and non-enhancing tumor, 2 for edema, and 4 for enhancing tumor). The segmentation targets are: enhancing tumor (ET, label 4), tumor core (TC, labels 1+4), and whole tumor (WT, labels 1+2+4).
\begin{table}[htbp]
	\centering
	
  \resizebox{\linewidth}{!}{
    \renewcommand\arraystretch{1.2}
        \small
	\begin{tabular}{c|c|cc|cc}
            \hline
		\multirow{2}{*}{Dataset} &\multirow{2}{*}{Method} & \multicolumn{2}{c}{Training Sets} & \multicolumn{2}{c}{Testing Sets} \\ 
		&& Training & Valuation & \multicolumn{2}{c}{Testing}\\ \hline
		\multirow{2}{*}{BraTS 2019}&Sequences Data & 2483 & 275 & \multicolumn{2}{c}{702}  \\ 
            &Slices Data & 37246 & 4139 & \multicolumn{2}{c}{10540}  \\  \hline

		\multirow{2}{*}{BraTS 2023}&Sequences Data & 11250 & 3750 & \multicolumn{2}{c}{3763}  \\ 
            &Slices Data & 116250 & 38750 & \multicolumn{2}{c}{38905}  \\  \hline
        
    \noalign{\smallskip}
	\end{tabular}
  }
     \caption{
		  Data Number on BraTS 2019 and BraTS 2023 Datasets.
	}
	\label{table:dataset num}
\end{table}

To mitigate data imbalance caused by non-informative black backgrounds, images were cropped to remove these areas. Each processed image has a size of \(155 \times 160 \times 160\) and is subsequently divided into 155 two-dimensional slices of \(160 \times 160\). All models are trained on the same training set and evaluated on the same test set. Sequential segmentation models take either individual 2D slices as input or concatenated slice sequences of a specified length. Data distribution details are provided in Table~\ref{table:dataset num}. Additionally, Z-score normalization is applied separately to the foreground region of each MRI modality to account for contrast differences.

\subsection{Metrics and Implementation Details}
Our network is implemented using the PyTorch framework on Ubuntu 22.04, with all experiments running on an NVIDIA RTX 2080Ti. In the experiments, computations related to computational cost and inference time are all based on this device. The performance of all algorithms is evaluated using Dice Score and Hausdorff95 Distance.

Dice Score measures the similarity between two samples, particularly in medical image segmentation, assessing the consistency between predicted and ground truth segmentations. The calculation is as follows:
\begin{align}
\text{Dice} = \frac{2TP}{FP + 2TP + FN}
\end{align}
where \(TP\) (True Positives) represents the correctly predicted positive samples, \(FP\) (False Positives) represents the incorrectly predicted positive samples, and \(FN\) (False Negatives) represents the missed positive samples.

Hausdorff95 is a metric that measures the distance between two sets, specifically the average minimum distance from each element in one set to the closest element in the other set. In the context of image segmentation, it measures the spatial difference between the predicted and ground truth segmentation. The calculation is as follows:
\begin{equation}
\begin{aligned}
\text{Haus}(A,B) = \\\max \left( \max_{S_A \in S(A)} d(S_A, S(B)), \max_{S_B \in S(B)} d(S_B, S(A)) \right)
\end{aligned}
\end{equation}
where \(A\) and \(B\) are the two segmentation sets, \(d\) is the distance from an element to the closest point, and \(S(A)\) and \(S(B)\) represent the elements in \(A\) and \(B\), respectively.

\subsection{Ablation Study}
\begin{figure}[ht]
	\centering
	\includegraphics[width=0.48\textwidth]{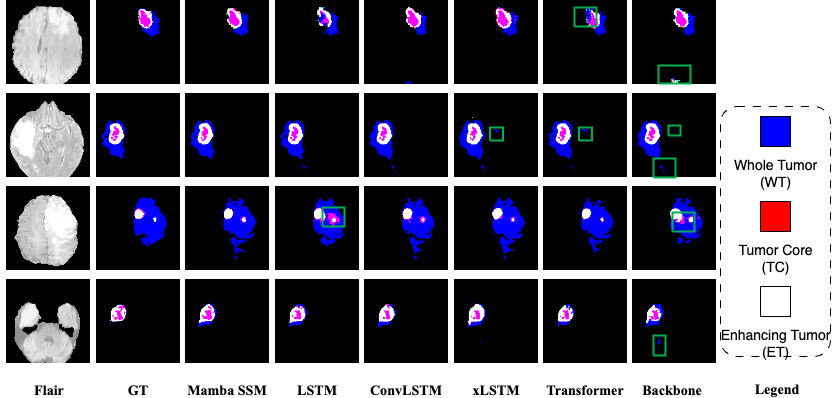} 
	\caption{Examples of Multi-sequential Module (TPS) segmentation results in the ablation study. From left to right: Flair modality input image, Ground Truth (GT), and segmentation results of different M-Net configurations.}
	\label{fig:abl}
\end{figure}

To evaluate the impact of different M-Net configurations and modules, Table~\ref{table:ablation study} presents the comparative results of various sequential models on the BraTS 2019 dataset. Compared to the backbone network without any sequential module and trained solely with single-slice (Slices) input—equivalent to the first phase of the TPS training strategy—all M-Net models incorporating the Mesh-Cast Sequential Module exhibit significant performance improvements. The Dice scores for the WT, TC, and ET regions improve by up to 1.2\%. Furthermore, compared to models trained with slice input only, all M-Net variants equipped with a sequential module achieve additional performance gains when trained with sequence input using the TPS strategy. Figure~\ref{fig:abl} also presents visualization results under different configurations, further validating the consistency of the performance improvements. Among all configurations, M-Net with Mamba SSM achieves the best performance with minimal additional FLOPs.

\setlength{\tabcolsep}{3pt}
\begin{table}[ht]
    \centering
	\renewcommand\arraystretch{1.2}
    \footnotesize
    \resizebox{0.47\textwidth}{!}{
    \begin{tabular}{c|c|ccc|ccc}
    \hline
        \multirow{2}{*}{Module and Method}& \multirow{2}{*}{FLOPs↓} & \multicolumn{3}{c}{Dice\_score(\%)} & \multicolumn{3}{c}{Hausdorff95} \\ \cline{3-8} 
		&& WT↑ & TC↑ & ET↑ & WT↓ & TC↓ & ET↓ \\ \hline
        Backbone(Slices) & 72.44G & 87.17 & 89.29 & 90.41 & 1.3710 & 0.8875 & 0.7093  \\ \hline
        Transformer(Slices) &\multirow{2}{*}{97.45G} & 87.24 & 89.30 & 90.29 & 1.3641 & 0.8791 & 0.6983 \\ 
        Transformer(TPS) && 87.56 & 89.96 & 90.79 & 1.3270 & 0.8354 & 0.6776  \\\hline
        LSTM(Slices) &\multirow{2}{*}{106.65G}& 87.59 & 89.78 & 90.56 & 1.3059 & 0.8454 & 0.6775  \\
        LSTM(TPS) && 88.06 & 89.97 & 90.73 & 1.2968 & 0.8340 & 0.6701  \\ \hline
        ConvLSTM(Slices) &\multirow{2}{*}{132.31G}& 87.74 & 89.92 & 90.68 & \textbf{1.3290 }& \textbf{0.8480} & 0.6905 \\ 
        ConvLSTM(TPS) && 88.19 & \textbf{90.22} & 90.79 & 1.3071 & 0.8358 & 0.6883 \\ \hline
        xLSTM(Slices) &\multirow{2}{*}{93.56G}& 87.92 & 89.60 & 90.77 & 1.3090 & 0.8707 & 0.6717 \\ 
        xLSTM(TPS) && \textbf{88.19} & 90.00 & \textbf{90.93} & 1.3040 & 0.8552 & \textbf{0.6689} \\ \hline
        Mamba SSM(Slices) &\multirow{2}{*}{91.29G}& 88.05 & 90.21 & 90.65 & 1.3332 & 0.8465 & 0.7064 \\ 
        Mamba SSM(TPS) && \color{red}\textbf{88.38} & \color{red}\textbf{90.52} & \color{red}\textbf{91.43 }& \color{red}\textbf{1.2869} & \color{red}\textbf{0.8154} & \color{red}\textbf{0.6571} \\ \hline
    \end{tabular}
    }
    \caption{
		Ablation Study of M-Net with Different Sequential Models on BraTS 2019 DATASET.
	}
	\label{table:ablation study}
\end{table}

In addition, we conduct an ablation study on the internal components of Mesh-Cast (see Table~\ref{tab:TPS_ablation}) to evaluate the individual contributions of temporal modeling (T) and channel modeling (C). Both components yield performance improvements when used independently, while their combination achieves the best results, confirming the effectiveness of modeling sequential information along both dimensions.

We also explored various configurations of the TPS (Two-Phase Scheduling) training strategy. The ablation results show that using only the first phase (randomized order) or only the second phase (ordered training) performs worse than the full two-phase strategy. In the first phase, shuffling the slice order improves the model’s generalization ability and helps prevent overfitting. In the second phase, the model learns the temporal dependencies between adjacent slices. Furthermore, the training order of shuffle-then-order outperforms the reverse (order-then-shuffle), further validating the rationality of our design.

\begin{table}[ht]
\centering
\resizebox{0.4\textwidth}{!}{
\begin{tabular}{l|ccc}
\toprule
\multirow{2}{*}{Model} & \multicolumn{3}{c}{Dice\_score(\%)}\\ \cline{2-4}
& WT & TC & ET \\
\midrule
Backbone (Ordered) & 87.17 & 89.29 & 90.41 \\
M-Net (T, Ordered) & 87.86 & 89.28 & 90.93 \\
M-Net (T+C, Ordered) & 88.05 & 90.21 & 90.65 \\ \hline
Backbone (Shuffled) & \textbf{88.21} & 90.11 & 90.86 \\
M-Net (T+C, Shuffled) & 88.07 & \textbf{90.32} & 91.05 \\
M-Net (T+C, Ordered+Shuffled) & 88.10 & 90.27 & \textbf{91.29} \\ \hline
M-Net (T+C, TPS) & \color{red}\textbf{88.38} & \color{red}\textbf{90.52} & \color{red}\textbf{91.43} \\
\bottomrule
\end{tabular}
}
\caption{Ablation study about TPS training strategy and Mesh-Cast Sequential Module on BraTS 2019 DATASET.}
\label{tab:TPS_ablation}
\end{table}

\begin{table*}[ht]
    \centering
    \renewcommand\arraystretch{1.3}
    \footnotesize
    \begin{tabular}{c|c|c|c|ccc|ccc}
    
    \hline
        \multirow{2}{*}{Model} & \multirow{2}{*}{Year} & \multirow{2}{*}{FLOPs↓} & \multirow{2}{*}{Inf Time(min)↓} & \multicolumn{3}{c}{Dice\_score(\%)} & \multicolumn{3}{c}{Hausdorff95} \\ \cline{5-10} 
		&&&& WT↑ & TC↑ & ET↑ & WT↓ & TC↓ & ET↓ \\ \hline
        UNet & 2015 & 321.19G & 12:32& 87.36/90.71 & 88.59/93.05 & 90.69/\textbf{93.36} & 1.3582/1.1863 & 0.9076/0.7329 & 0.6897/0.6730  \\ \hline
        SegResNet & 2019 & 5.98G & 10:54& 87.89/90.55 & 89.58/92.99 & 91.14/92.65 & 1.2977/1.1987 & 0.8403/0.7282 & 0.6649/0.7118 \\ \hline
        TransUNet & 2021 & 237.83G & 11:02&  84.50/90.71 & 86.72/92.52 & 88.39/92.92 & 1.3911/1.1810 & 0.9300/0.7276 & 0.7396/0.6869 \\ \hline
        nnUNet & 2021 & 82.00G & 97:67& 87.81/90.34 & \textbf{90.23}/92.74 & 90.96/92.37 & \textbf{1.2970}/1.2100 & 0.8311/0.7358 & 0.6628/0.6722 \\ \hline
        Transnorm & 2022 & 253.25G & 12:11& 86.56/87.97 & 87.88/91.82 & 89.28/91.49 & 1.3414/1.2226 & 0.8952/0.7299 & 0.7102/0.7247 \\ \hline
        UNETR & 2022 & 150.71 & 18:31& 85.29/88.35 & 87.16/89.16 & 89.54/91.43 & 1.3831/1.2427 & 0.9504/0.8926 & 0.7042/0.7211 \\ \hline
        Swin UNETR & 2022 & 136.80 & 21:33& 88.16/\textbf{91.11} & 88.85/93.20 & 90.86/\color{red}\textbf{93.42} & 1.3077/\textbf{1.1629} & 0.9119/0.7088 & 0.6814/\textbf{0.6631} \\ \hline
        MedNeXt & 2023 & 1.98G & 29:42& 87.55/89.91 & 89.18/92.82 & 90.45/92.85 & 1.3330/1.2160 & 0.8800/0.7303 & 0.6958/0.6953 \\ \hline
        SLf-UNet & 2024 & 534.73G & 17:26& 87.55/90.81 & 88.21/93.18 & 90.38/93.30 & 1.3273/1.1748 & 0.9032/0.7100 & 0.6871/0.6709\\ \hline
        MedSAM & 2024 & 166.55G & 30:19 & 85.39/88.55 & 87.90/91.55 & 88.20/90.30 & 1.4409/1.3155 & 0.9224/0.8003 & 0.7667/0.8153 \\ \hline
        Mamba UNet & 2024 & 72.44G & 14:12& \textbf{88.21}/91.03 & 90.11/\textbf{93.32} & 90.86/93.31 & 1.3061/1.1734 & \textbf{0.8235}/\color{red}\textbf{0.7008} & 0.6750/0.6764 \\ \hline
        UKAN & 2024 & 62.21G & 19:43& 87.39/90.64 & 89.50/93.04 & \textbf{91.20}/93.14 & 1.2989/1.1862 & 0.8415/0.7234 & \textbf{0.6585}/0.6824 \\ \hline
        M-Net & ours & 91.29G & 15:33& \color{red}\textbf{88.38}/\textbf{91.33}& \color{red}\textbf{90.52}/\textbf{93.55} & \color{red}\textbf{91.43}/\textbf{93.42}& \color{red}\textbf{1.2869}/\textbf{1.1534} & \color{red}\textbf{0.8154}\color{black}/\textbf{0.7069} & \color{red}\textbf{0.6571}/\textbf{0.6600} \\ \hline
    \end{tabular}
    \caption{
		Comparison with The SOTA Methods on BRATS 2019 and BraTS-2023 Datasets.
	}
	\label{table:com}
\end{table*}

\begin{figure*}[ht]
	\centering
	\includegraphics[width=1.0\textwidth]{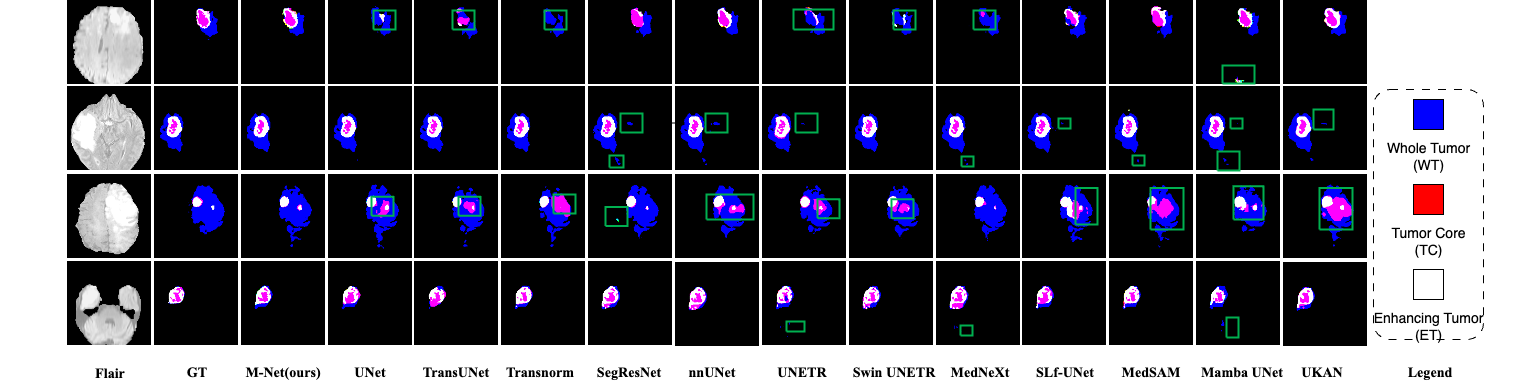} 
	\caption{Examples of segmentation results from multiple methods. From left to right: Flair modality input image, Ground Truth (GT), the proposed M-Net, and segmentation results from various comparison algorithms.}
	\label{fig:com}
\end{figure*}

On the other hand, TPS shows strong robustness under modality absence. Phase one uses the same input as standard 2D models, while phase two supports training on partial subsequences due to its flexible sequence design.

\subsection{Comparative Experiments}

We compared the proposed M-Net with several advanced methods, including U-Net\cite{Ron1}, TransUNet\cite{transunet}, nnUNet\cite{nnunet}, TransNorm\cite{transnorm}, UNETR\cite{unetr},  SegResNet\cite{segresnet}, SwinUNETR\cite{swin}, MedNeXt\cite{mednext}, UKAN\cite{UKAN}, SLf-UNet\cite{slf}, MedSAM\cite{MedSAM} and Mamba UNet\cite{mambaunet}. To ensure a fair comparison, all methods were trained on the same dataset using the same loss function, with a maximum of 300 training epochs and early stopping applied at 30 epochs. In the table, the best results are highlighted in bold red, and the second-best results are shown in bold black.

As shown in Table~\ref{table:com}, M-Net outperforms existing methods on nearly all regions and evaluation metrics across both the BraTS 2019 and BraTS 2023 datasets. On BraTS 2019, M-Net achieves Dice scores of 88.38, 90.52, and 91.43 for the WT, TC, and ET regions, respectively, with corresponding Hausdorff95 distances of 1.2869, 0.8154, and 0.6571. M-Net surpasses existing algorithms across all regions. Figure~\ref{fig:com} presents qualitative comparisons, where green boxes highlight typical detection errors. On the BraTS 2023 dataset, M-Net further achieves Dice scores of 91.33, 93.55, and 93.42 for the WT, TC, and ET regions, respectively, demonstrating its continued superiority.

Moreover, M-Net achieves excellent performance with low computational cost on BraTS 2023. It requires only 15 minutes for inference, compared to 97 minutes for nnUNet—just 15\% of the time.

These results show that M-Net offers strong segmentation across regions, balancing accuracy, speed, and efficiency via its 2D sequence-based design, confirming the viability of sequence-based MRI tumor segmentation.

\section{Conclusion}
Accurate segmentation of brain tumor MRI images is critical for diagnosis and treatment. Given the structural and spatial consistency of brain tumor MRI slices, we address this challenge by exploiting the inherent sequential nature of MRI slices through M-Net, leveraging the proposed Mesh-Cast mechanism to capture key “temporal-like” spatial correlations between consecutive slices. Our Two-Phase Sequential training strategy enhances the model's ability to learn both common anatomical patterns and specific sequential characteristics. Experiments on BraTS2019 and BraTS2023 datasets demonstrate that M-Net achieves superior segmentation performance while maintaining computational efficiency and inference time. The proposed techniques extend beyond MRI segmentation to other visual tasks with consistent data patterns, establishing a general spatiotemporal-aware sequential analysis paradigm. Future work will explore applications in multi-modal medical imaging and other domains with sequential structures. 

\section*{Acknowledgments}
This work was supported by Beijing Natural Science Foundation of China (QY24306, 4242034), the National Natural Science Foundation of China (62476178), and the National Key Laboratory of Human-Machine Hybrid Augmented Intelligence, Xi'an Jiaotong University (No. HMHAI-202407).

{
    \small
    \balance

    \bibliographystyle{ieeenat_fullname}
    \bibliography{main}

\begin{thebibliography}{38}
\providecommand{\natexlab}[1]{#1}
\providecommand{\url}[1]{\texttt{#1}}
\expandafter\ifx\csname urlstyle\endcsname\relax
  \providecommand{\doi}[1]{doi: #1}\else
  \providecommand{\doi}{doi: \begingroup \urlstyle{rm}\Url}\fi

\bibitem[AboElenein et~al.(2022)AboElenein, Piao, Noor, and Ahmed]{mirau}
N.~M. AboElenein, S. Piao, A. Noor, and P.~N. Ahmed.
\newblock Mirau-net: An improved neural network based on u-net for gliomas segmentation.
\newblock \emph{Signal Processing: Image Communication}, 101:\penalty0 116553, 2022.

\bibitem[Azad et~al.(2022)Azad, Al-Antary, Heidari, and Merhof]{transnorm}
R. Azad, M.~T. Al-Antary, M. Heidari, and D. Merhof.
\newblock Transnorm: Transformer provides a strong spatial normalization mechanism for a deep segmentation model.
\newblock \emph{IEEE Access}, 10:\penalty0 108205--108215, 2022.

\bibitem[Bakas et~al.(2017)Bakas, Akbari, Sotiras, Bilello, Rozycki, Kirby, and Davatzikos]{data2}
S. Bakas, H. Akbari, A. Sotiras, M. Bilello, M. Rozycki, J.~S. Kirby, and C. Davatzikos.
\newblock Advancing the cancer genome atlas glioma mri collections with expert segmentation labels and radiomic features.
\newblock \emph{Scientific data}, 4\penalty0 (1):\penalty0 1--13, 2017.

\bibitem[Bakas et~al.(2018)Bakas, Reyes, Jakab, Bauer, Rempfler, Crimi, and Jambawalikar]{data3}
S. Bakas, M. Reyes, A. Jakab, S. Bauer, M. Rempfler, A. Crimi, and S.~R. Jambawalikar.
\newblock Identifying the best machine learning algorithms for brain tumor segmentation, progression assessment, and overall survival prediction in the brats challenge.
\newblock \emph{arXiv preprint}, 2018.

\bibitem[Beck et~al.(2024)Beck, P{\"o}ppel, Spanring, Auer, Prudnikova, Kopp, and Hochreiter]{xlstm}
M. Beck, K. P{\"o}ppel, M. Spanring, A. Auer, O. Prudnikova, M. Kopp, and S. Hochreiter.
\newblock xlstm: Extended long short-term memory.
\newblock \emph{arXiv preprint}, 2024.

\bibitem[Beers et~al.(2017)Beers, Chang, Brown, Sartor, Mammen, Gerstner, and Kalpathy-Cramer]{Beers}
A. Beers, K. Chang, J. Brown, E. Sartor, C.~P. Mammen, E. Gerstner, and J. Kalpathy-Cramer.
\newblock Sequential 3d u-nets for biologically-informed brain tumor segmentation.
\newblock \emph{arXiv preprint arXiv:1709.02967}, 2017.

\bibitem[Chen et~al.(2021)Chen, Lu, Yu, Luo, Adeli, Wang, and Zhou]{transunet}
J. Chen, Y. Lu, Q. Yu, X. Luo, E. Adeli, Y. Wang, and Y. Zhou.
\newblock Transunet: Transformers make strong encoders for medical image segmentation.
\newblock \emph{arXiv preprint arXiv:2102.04306}, 2021.

\bibitem[Dandıl and Karaca(2021)]{Dan1}
E. Dandıl and S. Karaca.
\newblock Detection of pseudo brain tumors via stacked lstm neural networks using mr spectroscopy signals.
\newblock \emph{Biocybernetics and Biomedical Engineering}, 41\penalty0 (1):\penalty0 173--195, 2021.

\bibitem[Ding et~al.(2023)Ding, Lu, Cai, Zhang, and Shang]{slf}
H. Ding, J. Lu, J. Cai, Y. Zhang, and Y. Shang.
\newblock Slf-unet: Improved unet for brain mri segmentation by combining spatial and low-frequency domain features.
\newblock In \emph{Computer Graphics International Conference}, pages 415--426. Cham: Springer Nature Switzerland, 2023.

\bibitem[Feng et~al.(2020)Feng, Tustison, Patel, and Meyer]{Feng}
X. Feng, N.~J. Tustison, S.~H. Patel, and C.~H. Meyer.
\newblock Brain tumor segmentation using an ensemble of 3d u-nets and overall survival prediction using radiomic features.
\newblock \emph{Frontiers in computational neuroscience}, 14:\penalty0 25, 2020.

\bibitem[Fukushima(1980)]{Fuku}
K. Fukushima.
\newblock Neocognitron: A self-organizing neural network model for a mechanism of pattern recognition unaffected by shift in position.
\newblock \emph{Biological cybernetics}, 36\penalty0 (4):\penalty0 193--202, 1980.

\bibitem[Graves(2012)]{lstm}
A. Graves.
\newblock \emph{Long short-term memory}.
\newblock 2012.

\bibitem[Gu and Dao(2023)]{mamba}
A. Gu and T. Dao.
\newblock Mamba: Linear-time sequence modeling with selective state spaces.
\newblock \emph{arXiv preprint arXiv:2312.00752}, 2023.

\bibitem[Gu et~al.(2020)Gu, Wang, Song, Huang, Aertsen, Deprest, and Zhang]{canet}
R. Gu, G. Wang, T. Song, R. Huang, M. Aertsen, J. Deprest, and S. Zhang.
\newblock Ca-net: Comprehensive attention convolutional neural networks for explainable medical image segmentation.
\newblock \emph{IEEE transactions on medical imaging}, 40\penalty0 (2):\penalty0 699--711, 2020.

\bibitem[Han et~al.(2022)Han, Wang, Chen, Chen, Guo, Liu, and Tao]{Han1}
K. Han, Y. Wang, H. Chen, X. Chen, J. Guo, Z. Liu, and D. Tao.
\newblock A survey on vision transformer.
\newblock \emph{IEEE transactions on pattern analysis and machine intelligence}, 45\penalty0 (1):\penalty0 87--110, 2022.

\bibitem[Hatamizadeh et~al.(2021)Hatamizadeh, Nath, Tang, Yang, Roth, and Xu]{swin}
A. Hatamizadeh, V. Nath, Y. Tang, D. Yang, H.~R. Roth, and D. Xu.
\newblock Swin unetr: Swin transformers for semantic segmentation of brain tumors in mri images.
\newblock In \emph{International MICCAI brainlesion workshop}, pages 272--284. Cham: Springer International Publishing, 2021.

\bibitem[Hatamizadeh et~al.(2022)Hatamizadeh, Tang, Nath, Yang, Myronenko, Landman, and Xu]{unetr}
A. Hatamizadeh, Y. Tang, V. Nath, D. Yang, A. Myronenko, B. Landman, and D. Xu.
\newblock Unetr: Transformers for 3d medical image segmentation.
\newblock In \emph{Proceedings of the IEEE/CVF winter conference on applications of computer vision}, pages 574--584, 2022.

\bibitem[Isensee et~al.(2021)Isensee, Jaeger, Kohl, and et~al.]{nnunet}
Fabian Isensee, Paul~F. Jaeger, Simon A.~A. Kohl, and et al.
\newblock nnu-net: a self-configuring method for deep learning-based biomedical image segmentation.
\newblock \emph{Nature Methods}, 18\penalty0 (2):\penalty0 203--211, 2021.

\bibitem[Ji et~al.(2022)Ji, Xiao, Chou, and et~al.]{video2}
Guoping Ji, Guoli Xiao, Yung-Chieh Chou, and et al.
\newblock Video polyp segmentation: A deep learning perspective.
\newblock \emph{Machine Intelligence Research}, 19\penalty0 (6):\penalty0 531--549, 2022.

\bibitem[Li et~al.(2024)Li, Liu, Li, Wang, Liu, and Yuan]{UKAN}
C. Li, X. Liu, W. Li, C. Wang, H. Liu, and Y. Yuan.
\newblock U-kan makes strong backbone for medical image segmentation and generation.
\newblock \emph{arXiv preprint}, 2024.

\bibitem[Li et~al.(2019)Li, Sun, Meng, Liang, Wu, and Li]{diceloss}
X. Li, X. Sun, Y. Meng, J. Liang, F. Wu, and J. Li.
\newblock Dice loss for data-imbalanced nlp tasks.
\newblock \emph{arXiv preprint}, 2019.

\bibitem[Liang and Lauterbur(2000)]{Liang1}
Z.~P. Liang and P.~C. Lauterbur.
\newblock \emph{Principles of magnetic resonance imaging}.
\newblock SPIE Optical Engineering Press, Bellingham, 2000.

\bibitem[Liu et~al.(2024{\natexlab{a}})Liu, Tian, Zhao, et~al.]{liu2024vmamba}
Yujun Liu, Yuxin Tian, Yujie Zhao, et~al.
\newblock Vmamba: Visual state space model.
\newblock \emph{Advances in Neural Information Processing Systems}, 37:\penalty0 103031--103063, 2024{\natexlab{a}}.

\bibitem[Liu et~al.(2024{\natexlab{b}})Liu, Wang, Vaidya, Ruehle, Halverson, Soljačić, and Tegmark]{kan}
Z. Liu, Y. Wang, S. Vaidya, F. Ruehle, J. Halverson, M. Soljačić, and M. Tegmark.
\newblock Kan: Kolmogorov-arnold networks.
\newblock \emph{arXiv preprint}, 2024{\natexlab{b}}.

\bibitem[Mehta and Arbel(2018)]{Mehta}
R. Mehta and T. Arbel.
\newblock 3d u-net for brain tumour segmentation.
\newblock In \emph{International MICCAI Brainlesion Workshop}, pages 254--266. Cham: Springer International Publishing, 2018.

\bibitem[Menze et~al.(2014)Menze, Jakab, Bauer, Kalpathy-Cramer, Farahani, Kirby, and Van~Leemput]{data1}
B.~H. Menze, A. Jakab, S. Bauer, J. Kalpathy-Cramer, K. Farahani, J. Kirby, and K. Van~Leemput.
\newblock The multimodal brain tumor image segmentation benchmark (brats).
\newblock \emph{IEEE transactions on medical imaging}, 34\penalty0 (10):\penalty0 1993--2024, 2014.

\bibitem[Myronenko(2019)]{segresnet}
A. Myronenko.
\newblock 3d mri brain tumor segmentation using autoencoder regularization.
\newblock In \emph{Brainlesion: Glioma, Multiple Sclerosis, Stroke and Traumatic Brain Injuries: 4th International Workshop, BrainLes 2018, Held in Conjunction with MICCAI 2018, Granada, Spain, September 16, 2018, Revised Selected Papers, Part II 4}, pages 311--320. Springer International Publishing, 2019.

\bibitem[Pfeuffer et~al.(2019)Pfeuffer, Schulz, and Dietmayer]{Pfe}
A. Pfeuffer, K. Schulz, and K. Dietmayer.
\newblock Semantic segmentation of video sequences with convolutional lstms.
\newblock In \emph{2019 IEEE intelligent vehicles symposium (IV)}, pages 1441--1447. IEEE, 2019.

\bibitem[Ronneberger et~al.(2015)Ronneberger, Fischer, and Brox]{Ron1}
O. Ronneberger, P. Fischer, and T. Brox.
\newblock U-net: Convolutional networks for biomedical image segmentation.
\newblock In \emph{Medical image computing and computer-assisted intervention–MICCAI 2015: 18th international conference, Munich, Germany, October 5-9, 2015, proceedings, part III 18}, pages 234--241. Springer International Publishing, 2015.

\bibitem[Roy et~al.(2023)Roy, Koehler, Ulrich, Baumgartner, Petersen, Isensee, and Maier-Hein]{mednext}
S. Roy, G. Koehler, C. Ulrich, M. Baumgartner, J. Petersen, F. Isensee, and K.~H. Maier-Hein.
\newblock Mednext: transformer-driven scaling of convnets for medical image segmentation.
\newblock In \emph{International Conference on Medical Image Computing and Computer-Assisted Intervention}, pages 405--415, Cham, 2023. Springer Nature Switzerland.

\bibitem[Shahzadi et~al.(2018)Shahzadi, Tang, Meriadeau, and Quyyum]{Sha1}
I. Shahzadi, T.~B. Tang, F. Meriadeau, and A. Quyyum.
\newblock Cnn-lstm: Cascaded framework for brain tumour classification.
\newblock In \emph{2018 IEEE-EMBS Conference on Biomedical Engineering and Sciences (IECBES)}, pages 633--637. IEEE, 2018.

\bibitem[Tan et~al.(2020)Tan, Ashley, López, Malinzak, Friedman, and Khasraw]{Tan1}
A.~C. Tan, D.~M. Ashley, G.~Y. López, M. Malinzak, H.~S. Friedman, and M. Khasraw.
\newblock Management of glioblastoma: State of the art and future directions.
\newblock \emph{CA: a cancer journal for clinicians}, 70\penalty0 (4):\penalty0 299--312, 2020.

\bibitem[Vaswani(2017)]{transformer}
A. Vaswani.
\newblock Attention is all you need.
\newblock In \emph{Advances in Neural Information Processing Systems}, 2017.

\bibitem[Wang et~al.(2024)Wang, Zheng, Zhang, Cui, and Li]{mambaunet}
Z. Wang, J.~Q. Zheng, Y. Zhang, G. Cui, and L. Li.
\newblock Mamba-unet: Unet-like pure visual mamba for medical image segmentation.
\newblock \emph{arXiv preprint}, 2024.

\bibitem[Wu et~al.(2023)Wu, Ji, Liu, Fu, Xu, Xu, and Jin]{MedSAM}
J. Wu, W. Ji, Y. Liu, H. Fu, M. Xu, Y. Xu, and Y. Jin.
\newblock Medical sam adapter: Adapting segment anything model for medical image segmentation.
\newblock \emph{arXiv preprint arXiv:2304.12620}, 2023.

\bibitem[Xu et~al.(2019)Xu, Ma, Sun, Wu, Liu, and Kong]{Xu1}
F. Xu, H. Ma, J. Sun, R. Wu, X. Liu, and Y. Kong.
\newblock Lstm multi-modal unet for brain tumor segmentation.
\newblock In \emph{2019 IEEE 4th international conference on image, vision and computing (ICIVC)}, pages 236--240. IEEE, 2019.

\bibitem[Zhang et~al.(2020)Zhang, Xie, Wang, and Xia]{Zhang1}
J. Zhang, Y. Xie, Y. Wang, and Y. Xia.
\newblock Inter-slice context residual learning for 3d medical image segmentation.
\newblock \emph{IEEE Transactions on Medical Imaging}, 40\penalty0 (2):\penalty0 661--672, 2020.

\bibitem[Zhou et~al.(2022)Zhou, Porikli, Crandall, and et~al.]{video1}
Tianfei Zhou, Fatih Porikli, David~J. Crandall, and et al.
\newblock A survey on deep learning technique for video segmentation.
\newblock \emph{IEEE Transactions on Pattern Analysis and Machine Intelligence}, 45\penalty0 (6):\penalty0 7099--7122, 2022.

\end{thebibliography}
}

\end{document}